\pdfoutput=1

\documentclass[11pt]{article}

\usepackage[]{acl}

\usepackage{times}
\usepackage{latexsym}

\usepackage[T1]{fontenc}

\usepackage[utf8]{inputenc}

\usepackage{microtype}

\usepackage{inconsolata}
\usepackage{fancyhdr}
\usepackage{setspace}
\usepackage{cleveref}
\usepackage{tikz}
\usepackage{booktabs}
\usepackage{tipa}
\usepackage{subcaption}
\usepackage{makecell}
\usepackage{CJKutf8}

\usepackage{gb4e}
\noautomath

\crefname{xnumi}{example}{examples}
\creflabelformat{xnumi}{(#2#1#3)}
\crefname{xnumii}{example}{examples}
\creflabelformat{xnumii}{(#2#1#3)}
\crefname{xnumiii}{example}{examples}
\creflabelformat{xnumiii}{(#2#1#3)}
\crefname{xnumiv}{example}{examples}
\creflabelformat{xnumiv}{(#2#1#3)}

\usepackage{soul}

\newcommand{\ignore}[1]{}

\newcommand\autocite{\citep} \newcommand\textcite{\citet}
 
\title{Encoding of lexical tone in self-supervised models of
  spoken language}

\author{
    Gaofei Shen$^{1}$ ~ Michaela Watkins$^{2}$  ~ Afra Alishahi$^{1}$ 
    ~ Arianna Bisazza$^{3}$ ~ Grzegorz Chrupa\l{}a$^1$ \\
    $^1$ Tilburg University ~ $^2$ University of Amsterdam ~ $^3$ University of Groningen\\
    \texttt{\{g.shen, a.alishahi\}@tilburguniversity.edu}\\
    \texttt{m.m.watkins@uva.nl} ~ \texttt{a.bisazza@rug.nl} \\
    \texttt{grzegorz@chrupala.me}\\
    }

\begin{document}
\maketitle

\begin{abstract}
  Interpretability research has shown that self-supervised Spoken Language
  Models (SLMs) encode a wide variety of features in human speech from the
  acoustic, phonetic, phonological, syntactic and semantic levels, to speaker
  characteristics. The bulk of prior research on representations of phonology
  has focused on segmental features such as phonemes;  the encoding of
  suprasegmental phonology (such as tone and stress patterns) in SLMs is not yet
  well understood. Tone is a suprasegmental feature that is present in more than
  half of the world's languages. This paper aims to analyze the tone encoding
  capabilities of SLMs, using Mandarin and Vietnamese as case studies. We show
  that SLMs encode lexical tone to a significant degree even when they are
  trained on data from non-tonal languages. We further find that SLMs behave
  similarly to native and non-native human participants in tone and consonant
  perception studies, but they do not follow the same developmental trajectory.
\end{abstract}

\section{Introduction}
Explaining the inner workings of self-supervised models of written and spoken
language has been the focus of much recent work. Transformer-based
\autocite{vaswaniAttentionAllYou2017} written language models have been shown to
encode many types of linguistic information
\autocite{conneauWhatYouCan2018,hewittStructuralProbeFinding2019}. The analysis
of self-supervised Spoken Language Models (SLMs) is also gaining traction:
architectures such as wav2vec2
\autocite{baevskiWav2vecFrameworkSelfSupervised2020} and HuBERT
\autocite{hsuHuBERTSelfSupervisedSpeech2021} have been shown to encode
linguistic information at the phonetic, phonological, syntactic and semantic
levels of human speech without labeled data
\autocite{abdullahHowFamiliarDoes2021,maProbingAcousticRepresentations2021,deSeyssel2022ProbingPL,barteldsNeuralRepresentationsModeling2022,martinProbingSelfsupervisedSpeech2023,
shenWaveSyntaxProbing2023, pasadWhatSelfSupervisedSpeech2024}.

The majority of research on representations of phonetic and phonological
information in SLMs focuses on the segmental level. \textbf{Segmental} refers to
units of speech that do not spread but remain localized. Phonemes (e.g.\ vowels
and consonants) are the smallest abstract units of sound that help to
distinguish one unit from another (e.g.\ \textbf{p}at vs \textbf{b}at).
\textbf{Suprasegmental}, in contrast, refers to features that are not
necessarily limited to single units, but can spread across multiple phonemes or
phrases. Examples include tone, stress patterns, and intonation, which can all
entail syllable and phrase level changes \cite{singhNewViewLanguage2016}. The
representation of suprasegmental information in SLMs is important to study, as
it is one of the main distinguishing features of speech compared to text: spoken
utterances use suprasegmental cues to convey information that is sometimes not
explicitly marked in a corresponding written sentence.\footnote{Mandarin does not
explicitly mark tone in writing, but Vietnamese does.} As a first step, in this
work, we focus on lexical tone as a highly constrained, relatively
well-understood example of a suprasegmental feature.

We firstly examine to what extent SLMs trained on tonal and non-tonal
languages encode tone information in their internal representations.
We find that SLMs are capable of capturing tonal information,
regardless of whether they are trained on tonal or non-tonal
languages.

Secondly, we investigate the impact of supervised fine-tuning on the
automatic speech recognition (ASR) task. 
We find that fine-tuning \textit{enhances} tone representations
for models trained on tonal languages, but \textit{reduces} them
for models trained on non-tonal languages.

Thirdly, we explore whether SLMs exhibit the same perceptual patterns as
native and non-native human listeners. We find that models show patterns similar
to humans in discrimination of Mandarin tones and consonants, but find no
evidence that they follow a similar developmental trajectory.

\section{Tones} \label{sec:Tones}

Estimates suggest that more than 60\% of the world's languages use some degree
of tonal contrast \autocite{yipTone2002}. The primary focus of this work is on lexical
tone, the process by which lexical items are distinguished from one another
primarily by pitch cues \autocite{chenComputationalModellingTone2022}. We focus
primarily on pitch cues: while there are suggestions that articulatory
cues alone are not sufficient for tonal distinction in languages such as Vietnamese, with
socio-linguistic register also being important if present in the language in
question \cite{Brunelle2009ToneVietnamese}, this is still subject to further
research in perception and production studies to validate as the body of work
remains small, particularly regarding perception. We therefore state here that
non-tonal phonemic units (e.g.\ vowels, consonants) can be defined primarily by
\textbf{non-pitch articulatory cues}, such as vowel height, voicing, and
duration. In contrast, tonal units make use of \textbf{pitch cues}, with F0
(fundamental frequency) contour usually considered to be the primary cue
\autocite{rheeGoingF0Acquisition2021}. In ambiguous contexts, other pitch cues
can be used in combination with non-pitch cues such as amplitude, voice quality
(e.g.\ breathy vs creaky), and spectral tilt
\autocite{rheeGoingF0Acquisition2021}. 

The Tone Bearing Unit (TBU) is the segment containing tone; this is typically,
although not always, found on the syllable level. In the case of the primary
language investigated in this paper, Mandarin, syllables are morpheme based (as in,
Mandarin syllables \textit{are} morphemes). Thus in Mandarin every morpheme
contains obligatory tone, and is in of itself the TBU \autocite{junAsianPacificRim2020}. 

\begin{figure}
    \centering
    \includegraphics[width = \linewidth]{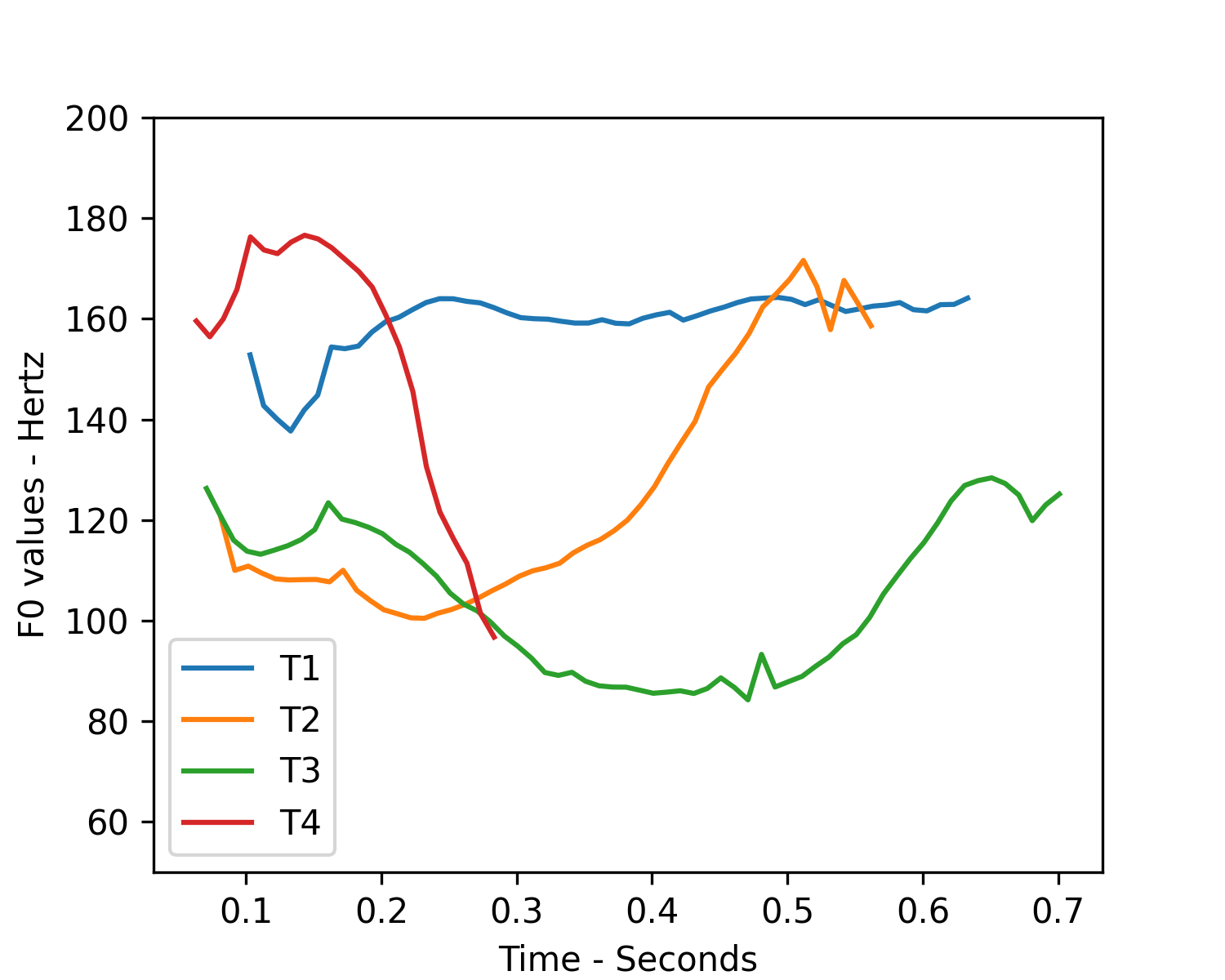}
    \caption{F0 contours of the four Mandarin tones measured from  pronunciations recorded by one of
    the co-authors, a native speaker of Mandarin Chinese.
    The four syllables are pronounced in isolation
    (notation: m\={a} T1,
    m\'{a} T2,
    m\v{a} T3, 
    m\`{a} T4).}
    \label{fig:toneIllustration}
\end{figure}

We compare SLMs trained on non-tonal languages as well as three fully lexical tonal
languages: Mandarin, Cantonese and Vietnamese. The models are tested
primarily on Mandarin data. Mandarin demonstrates full tonality,
with tone found on each morpheme
\autocite{hymanWhatToneTeaches2018}, and has been widely studied
for tone perception as well as acquisition. Secondarily, we also test on data from
another lexical tone language, Vietnamese, to assess if our results generalize.

\textbf{Mandarin Chinese} is typically described as containing four (lexical) tones and
one neutral tone that only occurs in unstressed syllables \autocite{Wu2020MandarinDuration}.
The tones are conventionally assigned the labels 1-4 (T1-4);
\Cref{fig:toneIllustration} illustrates the four Mandarin tones. 

Since one morpheme (one character) corresponds to one tone in Mandarin
Chinese, we can use the Pinyin transcription to obtain our tone labels easily
(see \Cref{fig:toneIllustration} for notations); for example:
\begin{CJK*}{UTF8}{gbsn}
    \begin{exe}
        \ex \label{ex:TBU}
        \glll 今 天 天 气 很 好 \\ 
            Jīn tiān tiān qì hěn hǎo \\ 
            T1 T1 T1 T4 T3 T3\\ 
        \glt `The weather today is very good.'
    \end{exe}
\end{CJK*}
The tone label corresponds to the tone of the character when it is pronounced in
isolation (base form). However, Mandarin features \textit{tone sandhi}, i.e.\ the tone
assigned to individual morphemes can change in pronunciation based on the tone
of the adjacent morpheme (sandhi form). One instance of tone sandhi rules in Mandarin is T3
sandhi \autocite{chenToneSandhiPatterns2000}: if two T3 (`dipping') tones occur
next to one another, the first will adjust to T2 (`rising') to avoid two
consecutive T3 tones, as can be seen in examples \ref{ex:toneSandhi1} and \ref{ex:toneSandhi2}, after
\textcite{chenToneSandhiPatterns2000}. Tone labels obtained from Pinyin
transcriptions only take the base form into account.

\begin{CJK*}{UTF8}{gbsn}
\begin{exe}
\ex \label{ex:toneSandhi1}
\begin{tabular}{ll}
     小 &\\
     xiǎo &\\
     T3 &\\
     `small' & \\
\end{tabular}
\end{exe}
\end{CJK*}
\begin{CJK*}{UTF8}{gbsn}
\begin{exe}
\ex \label{ex:toneSandhi2}
\begin{tabular}{lll}
     小 & 狗 & \\
     xiǎo& gǒu & \\
     T3& T3 & {\it base form}\\
     T2 &T3 & {\it sandhi form}\\
     `small & dog' & \\
\end{tabular}
\end{exe}
\end{CJK*}

The primary pitch cue that distinguishes the individual Mandarin tones from each other is
F0; however, secondary pitch cues are also present such as voice quality
and spectral tilt \cite{Belotel-Grenie1994PHONATIONCHINESE, huangDifferentAttributesCreaky2020}.  

\textbf{Vietnamese} also has obligatory tones on every syllable, similar to Mandarin's
morphemic TBU \autocite{kirbyVietnameseHanoiVietnamese2011}. We adhere to the
eight tone system described by \textcite{kirbyVietnameseHanoiVietnamese2011} in
our experiment setup.

\textbf{Cantonese} is a Sinitic language related to Mandarin, and also
features lexical tone, with six tonal distinctions \autocite{zeeChineseHongKong1991} as opposed to
Mandarin's four.

\section{Related work}

The present paper builds both on works interpreting the inner workings
of SLMs and on experiments on perception of aspects of human speech. 

\subsection{Analyzing SLMs}

The transformer architecture \autocite{vaswaniAttentionAllYou2017} has dominated the
SLM realm. Researchers have developed many methods to analyze the
inner-working of these models.
\textcite{pasadLayerwiseAnalysisSelfsupervised2021} provide an overview on the
variety of linguistic featurs encoded by self-supervised SLMs. The
models tend to
follow an autoencoder-like behavior with the middle layers showing the
strongest encoding of a variety of linguistic features.

More
recently, research has focused on specific properties of the input audio
that is being encoded by the models.
\textcite{martinProbingSelfsupervisedSpeech2023} tested whether SLMs
can distinguish between voiced and voiceless consonants.
\textcite{shenWaveSyntaxProbing2023} showed that self-supervised as
well as visually-supervised SLMs are capable
of encoding syntactic properties to some extent. Some prior works in
the field have touched on the encoding of suprasegmental features in SSL speech
models. \textcite{barteldsNeuralRepresentationsModeling2022} showed the hidden
state activations of SLMs are capable of capturing intonational and durational
information on the phrase level, indicating that they can encode non-segmental information to a
significant degree. 

Many recent interpretability studies are inspired by psycholinguistics and child
language development research. With the rise of probing and other
interpretability methods, researchers replicated experimental paradigms in
psychology and linguistics to better understand the capabilities of models
compared to humans. For example, \textcite{wilcoxUsingComputationalModels2023}
tested text language models using psycholinguistic experimental paradigms,
showing that they are capable of learning syntactic dependencies with relatively
little input data.

On the speech side, \textcite{lavechinStatisticalLearningModels2023}
presented evidence that self-supervised SLMs can develop limited
language-specific perception.
\textcite{cruzblandonIntroducingMetaanalysisEvaluation2023} proposed comparing
model behavior using checkpoints in the SLM pre-training process with
data in child language development. They showed that computational language
models can be a valuable resource in testing or confirming linguistic theories
in the language development field. The methodology is mostly concerned with the
overall learning of the language model in the output stage. Our work contributes
to the explanation of the inner workings of SLMs.

\subsection{Human perception experiments}

In terms of tone perception, F0 is a clear primary cue
\autocite{ryantMandarinToneClassification2014,rheeGoingF0Acquisition2021,chenComputationalModellingTone2022},
but other secondary pitch cues serve to assist when speech is ambiguous and/or
disrupted. Given that conversational speech contains non-trivial speech
recognition difficulties such as e.g.\ tone sandhi and coarticulation,
individual variation, and context omission
\cite{ryantMandarinToneClassification2014}, secondary cues
play a role in the distinction of tones. An example of this is voice
quality, where for example lowering F0 (introducing `creaky' voice) increased perceptual saliency for T3, whereas T1 and T4 accuracy decreased and T2 remained unaffected \autocite{huangDifferentAttributesCreaky2020, chaiSOURCECREAKMANDARIN2019, kuangCovariationVoiceQuality2017}. This emphasises the fact that F0 does
not operate in isolation, but that covariation between pitch and voice quality
is inherent in Mandarin. Spectral cues (e.g.\ amplitude differences, spectral
tilt) have also been suggested to be sufficient for adult speakers in tone
production, while children are thought to hyperarticulate the tonal differences
in speech \cite{rheeGoingF0Acquisition2021}.

Suprasegmental cues appear to be preferred in experiments that compare segmental
and suprasegmental cues against each other. Human infants are more sensitive to suprasegmental cues, with even newborns
showing the same preference
\autocite{mehlerPrecursorLanguageAcquisition1988,nazziLanguageDiscriminationNewborns1998}.
Several studies observe that tonal sensitivity develops earlier than
perception of vowels and
consonants \autocite{xiCategoricalPerceptionVOT2009,yeungWhenDoesNative2013},
with sensitivity to non-native tonal distinctions remaining longer
than perception of non-tonal non-native phoneme categories
\cite{liuPerceptionTonesInfants2014,shiPerceptionRepresentationLexical2017}.

Comparing vowels,
consonants and tones, \textcite{singhSpokenWordRecognition2015} show that
Mandarin learning children's sensitivity to consonants and vowels develop at a
similar rate and shows departure from tones. The effect of tone mispronunciation is
much larger than that of vowel or consonant mispronunciation for toddlers, but the 
pattern is reversed in preschoolers \autocite{singhSpokenWordRecognition2015}.

\subsection{Automatic classification of tones}

Automatic tone classification in Mandarin traditionally uses F0 contour and
mel-frequency cepstral coefficients (MFCC)
features. Advances in deep learning brought improvements in
performance of tone classification. 
\textcite{ryantHighlyAccurateMandarin2014} compare  MFCC features and
F0 contour as input to a neural tone classifier. MFCC
features, while not explicitly encoding the F0 contour information, achieve an error rate of 15.56\% for Tone 1-4 classification. The combination of MFCC features and F0 contours extracted with
different methods did not see an improvement in the classifier's performance,
indicating that the classifier was able to extract F0 contour from the MFCC
features, or it was able to predict Mandarin tones reliably
without F0 contour information. However, it is possible that the classifier was able to exploit associations
between specific phonemes strings and tone labels, and hence avoid learning to
detect tone based on pitch and voice quality cues.

After the introduction of self-supervised SLMs,
\textcite{yuanAutomaticRecognitionSuprasegmentals2021} fine-tuned an
English pre-trained wav2vec2 model
\autocite{baevskiWav2vecFrameworkSelfSupervised2020} for Mandarin tone classification and
achieved a tone error rate of 6\% on the same dataset as
\autocite{ryantHighlyAccurateMandarin2014}. Clearly, SLMs can handle
the task of classifying Mandarin lexical tone with labeled
fine-tuning. The aim of the present paper is not to compete with the
existing implementations of Mandarin tone classifiers; rather we aim
to uncover the tone encoding capabilities that emerge without explicit
supervision.

\section{Methodology} \label{sec:experiment1}
We use a number of wav2vec2-based
\autocite{baevskiWav2vecFrameworkSelfSupervised2020} models pre-trained and
fine-tuned on datasets of different languages for our investigation. As examples
of tonal languages, we choose Mandarin, Vietnamese and Cantonese, whereas
English and French serve as non-tonal language examples. The models
trained in the languages above are then tested on test data from Mandarin and Vietnamese.

To examine the encoding of tone, we train linear probing classifiers
on the hidden state activations extracted from the aforementioned models
for every morpheme in our testing datasets. \footnote{We release the codebase for our experiments at \url{https://github.com/techsword/tone-encoding-in-speech-model}}

\begin{table*}
  \centering
  \begin{tabular}{llrrc}
    \toprule
    &&\multicolumn{2}{c}{Size (hours)} & \\
    Training language &  Tonality&  Pre-training &
    Fine-tuning & Speech type\\ 
    \midrule
    English \autocite{baevskiWav2vecFrameworkSelfSupervised2020} &  Non-tonal&
    960 & 960 & Read\\ 
    French \autocite{parcolletLeBenchmarkStandardizedReplicable2023}&
    Non-tonal&  1,000 & - & Read\\ 
    \midrule
    Mandarin \autocite{luContextawareKnowledgeTransferring2022}&  Tonal&  1,000 &
    178 & Read \\ 
    Vietnamese \autocite{Thai_Binh_Nguyen_wav2vec2_vi_2021}&  Tonal&  13,000 &
    250 & YouTube audio/Read \\ 
    Cantonese  \autocite{huangWav2vecASRCantoneseSpeaking2023}&  Tonal&  2,800 &
    - & Spotaneous + Read\\ 
    \bottomrule
  \end{tabular}
  \caption{Description of the datasets used in pre-training/fine-tuning models.}
  \label{tab:trainingData}
\end{table*}


\subsection{Datasets}

\paragraph{Training data.} \label{sec:trainingData} 
We examine SLMs that were trained on datasets of the following languages: 

\textbf{Mandarin} pre-trained with AISHELL-2
\autocite{duAISHELL2TransformingMandarin2018} and fine-tuned with AISHELL-1
\autocite{buAISHELL1OpenSourceMandarin2017a}. \textbf{English} pre-trained and
fine-tuned with LibriSpeech \autocite{panayotovLibrispeechASRCorpus2015}.
\textbf{Vietnamese} pre-trained with unlabelled YouTube audio and fine-tuned
with the VLSP dataset for ASR \autocite{Thai_Binh_Nguyen_wav2vec2_vi_2021}.
\textbf{Cantonese} pre-trained on a combined dataset of older Cantonese adult
speech and YouTube audio \autocite{huangWav2vecASRCantoneseSpeaking2023}.
\textbf{French} pre-trained on MLS French
\autocite{pratapMLSLargeScaleMultilingual2020a}.
\noindent \Cref{tab:trainingData} summarizes the characteristics of these datasets.

\paragraph{Test data.} \label{sec:testData}
We primarily use the Mandarin Chinese THCHS-30 dataset
\autocite{wangTHCHS30FreeChinese2015} for testing models' encoding of Mandarin
tone.
THCHS-30 consists of 30 hours of Mandarin speech recorded in a laboratory
environment. The dataset is transcribed into both Chinese characters and
Mandarin Pinyin. We also obtain character-level forced alignment with the
Charsiu aligner \autocite{zhu2022charsiu}. 

To test the generalizability of our results, we also use the
Vietnamese VIVOS dataset
\autocite{luong-vu-2016-non}, which consists of 15 hours of Vietnamese read speech recorded in a
laboratory environment. The dataset is transcribed into Vietnamese orthography.
We then convert the transcription into International Phonetic
Alphabet (IPA) with tone labels with vPhon
\autocite{kirbyVPhonVietnamesePhonetizer2008}. We use the Montreal Forced
Aligner \autocite{mcauliffeMontrealForcedAligner2017} to obtain a syllable-level
forced alignment.

\paragraph{Pre-training data.} For the experiments on SLM's
learning trajectory and perceptual patterns (see \Cref{sec:human}), we pre-train SLMs from scratch on the following datasets:
\begin{itemize}
\item MAGICDATA \autocite{magicdata2019}, containing 755 hours of read Mandarin
Chinese. The dataset was pre-split into a 712-hour training set and a 28-hour
validation set. 
\item LibriSpeech \autocite{panayotovLibrispeechASRCorpus2015}, see details in \Cref{tab:trainingData}.
We split a
subset of the LibriSpeech dataset into a 710-hour training set and a 29-hour
validation set.
\end{itemize}
 

\subsection{Spoken Language Models} \label{sec:models}

\paragraph{Architecture.} With the exception of the Cantonese model, all models
investigated in this paper are based on the base configuration of wav2vec2
\autocite{baevskiWav2vecFrameworkSelfSupervised2020}. Wav2vec2-base consists of
five convolutional feature encoder and twelve transformer layers. The feature
encoder processes the audio waveform input into latent speech representations,
and the transformer layers encode the feature encoder output into contexual
representations. The wav2vec2-base models has 95M parameters. The Cantonese model uses the
wav2vec2-conformer architecture with 180M parameters.

\paragraph{Training objectives.} The fully self-supervised \emph{pre-training}
objective in wav2vec2 consists in 
discriminating between the matched and unmatched segment
representations for a masked portion of the latent speech representation. 
The ASR \emph{fine-tuning} objective consists in transcribing the audio input
into output tokens in the orthography of the target language and is realized by
adding a linear layer on top of a pre-trained wav2vec2 model.

\paragraph{Checkpoints.}
For the experiments in \Cref{sec:human} we pre-train two SLMs with the fairseq toolkit
\autocite{ottFairseqFastExtensible2019} on LibriSpeech for English and MAGICDATA
for Mandarin; we train both models for 85,000 steps using 8 Nvidia A100-40GB
GPU with update frequency = 8 to simulate training with 64 GPUs. Each model
finished training in approximately 96 hours. We save checkpoints every
5,000 steps.


\subsection{Probing classifiers} \label{sec:classTone} 

\paragraph{Preprocessing.} 
We follow previous work \autocite{ryantHighlyAccurateMandarin2014} in removing segments transcribed with the
neutral tone from the Mandarin tone classification task. Mandarin neutral tones
primarily appear in unstressed syllables (cf. \Cref{sec:Tones}) and
hence are more susceptible to variations.

\paragraph{Generating classifier input.}
We extract the hidden state activations of models
as a response to audio samples in the test data. We average-pool the hidden
state output corresponding to the duration of individual syllables
to obtain a vector using forced alignment timestamps. The resulting
768-dimensional vectors are input to the classifiers. 
To control for the influence of lexical cues on tone detection, 
we construct an exclusive train-test-split such that phoneme strings
appearing in the test set do not appear in the training set. This
setup prevents the probing classifier from exploiting associations
between tones and phoneme sequences. We employ a randomized 80:20 train-test
split with the split sizes shown in \Cref{tab:tonetraintestsplit}.
\begin{table}
  \centering
  \begin{tabular}{llr}
    \toprule
    Language & Split & Samples\\
    \midrule
    Mandarin & Train & 223,851\\
    Mandarin & Test & 45,772\\
    Vietnamese & Train &  124,248     \\
    Vietnamese & Test  &  29,629     \\
    \bottomrule
  \end{tabular}
  \caption{Train/test splits for the tone probing classifier, for the
    Mandarin and Vietnamese data. }
  \label{tab:tonetraintestsplit}
\end{table}

\paragraph{F0 and MFCC baselines.}
We closely follow \textcite{ryantHighlyAccurateMandarin2014} and use F0 contours
and 40-dimensional mel-frequency cepstral coefficients (MFCC) features as 
baselines. 
We use Librosa \autocite{mcfeeLibrosaLibrosa102023} to
extract the MFCC features and Praat
\autocite{jadoulIntroducingParselmouthPython2018,boersmaPraatDoingPhonetics2021}
to extract the F0 contours from the audio samples. We then find the center frame for
each word using the alignment timestamps and concatenate all frames in a 21
frame window (10-1-10) for both F0 and MFCC features. We end up with a 
21-dimensional vector for F0 contours and 840-dimensional vector for MFCC features
as our baseline classifier inputs.

\paragraph{Text baseline.}
In addition to audio baselines, we also include a text-based transformer model
in our comparison. BERT \autocite{devlinBERTPretrainingDeep2019} serves as a
reference point to show how much information is encoded in the speech
signal as opposed to what can be guessed from pure text. We use a
Chinese pre-trained BERT\footnote{\url{https://huggingface.co/bert-base-chinese}}  that encodes
Chinese characters into vectors. We extract per-word hidden state outputs with a resulting
768-dimensional vector.

\paragraph{Tone classifiers.}
We use the syllable activation vectors as input to a Ridge linear classifier
that predicts the lexical tone of the input morpheme. We select the final model
via 5-fold cross-validation, and report the classification accuracy on the test
split. The regularization strength $\alpha$ was tuned for values $\{ 10^n~|~ n
\in \{-4, -3, -2, -1, 0, 1, 2\} \}$.

\paragraph{Consonant classifiers.}
When comparing tone to consonant classification, we employ the same classifier
setup for consonant and replicate the perception experiment in
\textcite{wangAcquisitionMandarinConsonants2020} in \Cref{sec:trajectory}. Since
we only investigate consonants that appear solely in the onset position and the
rest of the phonemes are not relevant to our task, we use the same syllable
vectors as above instead of obtaining a phoneme vector with using phoneme level
alignment. We construct exclusive train-test-splits that contain unique rimes
(nucleus + coda) of the syllables. Specific details of the train/test split for
this experiment can be found in \Cref{tab:consonanttraintestsplit}.
\begin{table}
  \centering
  \begin{tabular}{lr}
    \toprule
    Split & Samples\\
    \midrule
    Train & 92,413\\
    Test & 15,688\\
    \bottomrule
  \end{tabular}
  \caption{Train/test split for the consonant probing classifier, for
    the Mandarin data. }
  \label{tab:consonanttraintestsplit}
\end{table}

\section{Results}
\label{sec:results}
In this section, we present a series of experiments for analyzing the encoding of tone in SLMs.

\subsection{Tone encoding across languages}
\label{sec:languages}

\begin{figure}[t]
  \includegraphics[width = \linewidth]{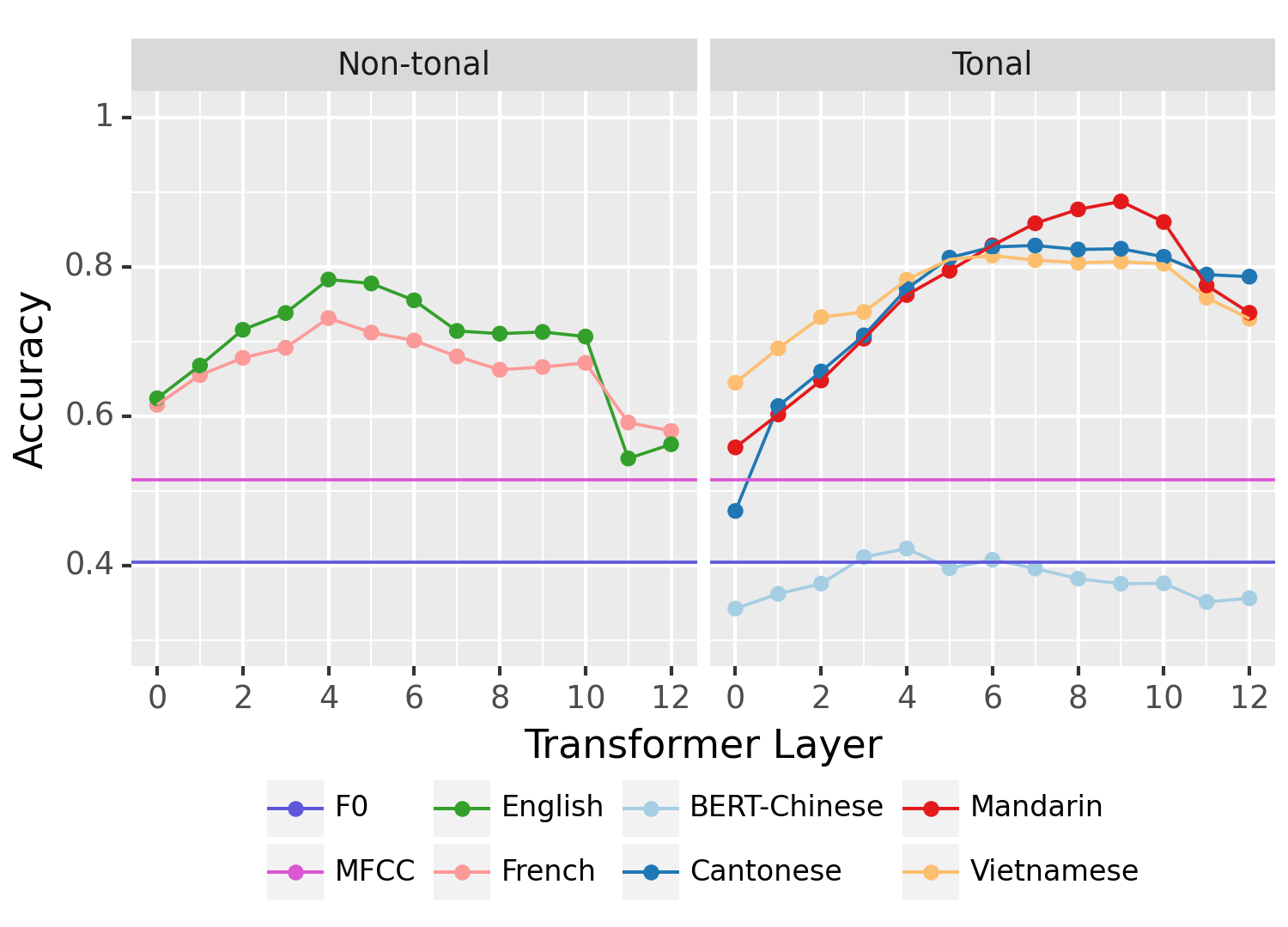}
  \caption{Classification accuracy of Mandarin lexical tones using layer-wise representations from models pre-trained on tonal and non-tonal languages.}
  \label{fig:languageTone}
  \includegraphics[width = \linewidth]{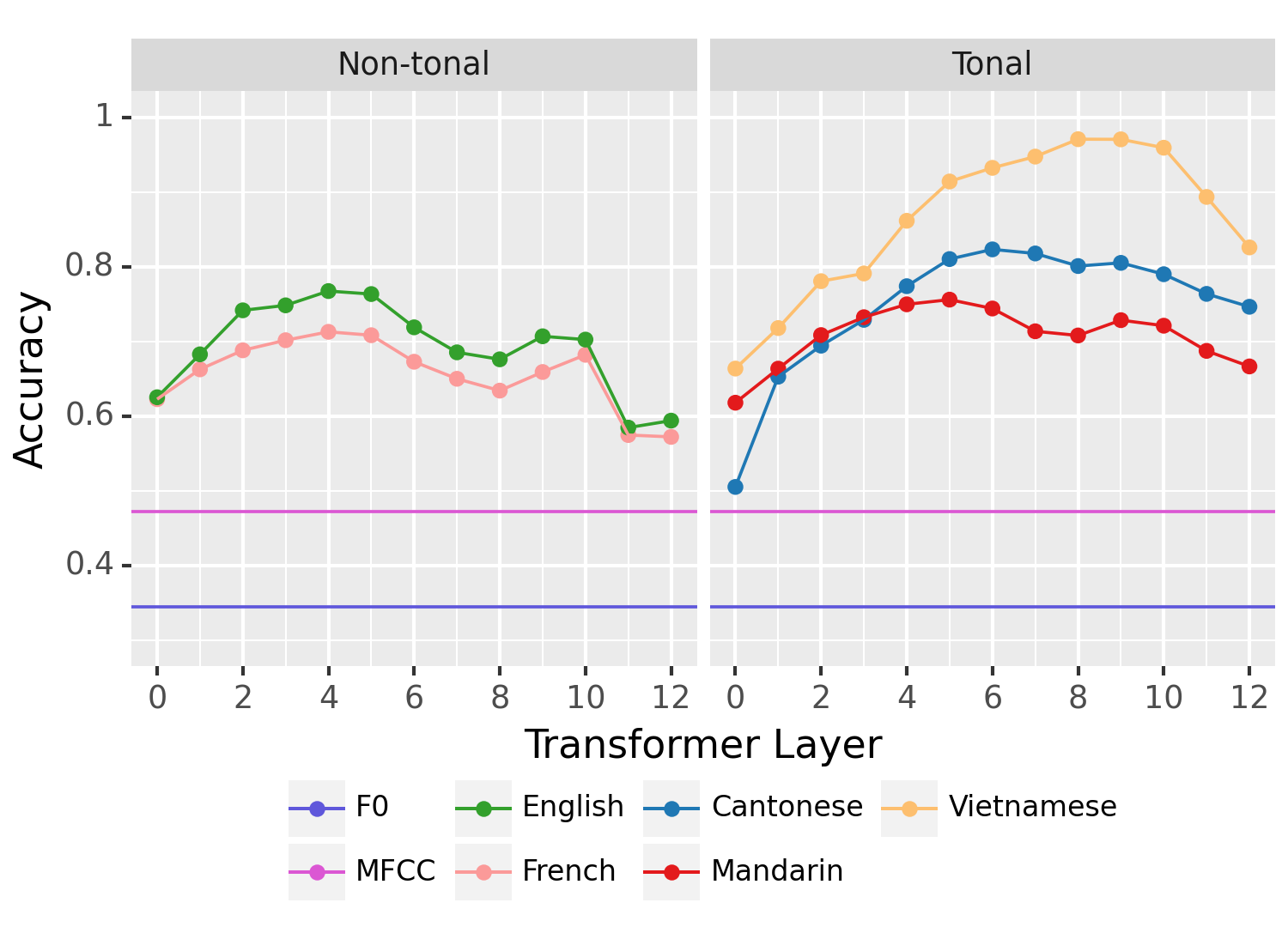}
  \caption{Classification accuracy of Vietnamese lexical tones with hidden-state activations from models pre-trained on tonal and non-tonal languages.}
  \label{fig:Vietnamese}
\end{figure}

\Cref{fig:languageTone} shows the tone classification accuracy using the
layer-wise representations of all models pre-trained on non-tonal (left) versus
tonal (right) languages. We see that all layers of all models perform better
than the F0 and MFCC baselines, which themselves outperform the text-based BERT
baseline. The classification accuracy for tonal language models is overall
higher, and increases in the higher layers of the models. Models trained on
non-tonal languages also show substantial encoding of tone; but remarkably,
there is a substantial drop in classification accuracy in their final layers
while the corresponding decrease is much less pronounced in tonal language
models. 

We repeat the tone classification experiment for Vietnamese tones. Results in
\Cref{fig:Vietnamese} show the Cantonese model performs slightly better than the
English model, especially towards the later layers; the Mandarin model, however,
patterns similar to the English model. This is could be due to the fact that
Mandarin has fewer tonal contrasts than Vietnamese and Cantonese (cf.
\Cref{sec:Tones}), or, more likely, that Vietnamese uses different acoustic cues
such as phonation type and voice quality in tonal perception than f0 contours
or height in Mandarin \autocite{Brunelle2009ToneVietnamese}. 

Studies on human participants show that speakers of other tonal languages can
perform better at identifying Mandarin lexical tones compared to non-tonal
language speakers \autocite{soCrosslanguagePerceptionNonnative2010}, but other
literature suggest that listeners are more sensitive to specific cues in tonal
perception that are present in their native languages
\autocite{dicanioCrosslinguisticPerceptionItunyoso2012,schaeferLexicalFunctionPitch2014}.
The SLMs we tested show the former pattern for Mandarin tone classification.
Regarding Vietnamese tones, the result is more equivocal suggesting that
Cantonese tone representations generalize to Vietnamese to some extent, while
Mandarin ones do not.


\subsection{Impact of ASR fine-tuning}
\label{sec:objective}

\begin{figure}[t]
    \includegraphics[width = \linewidth]{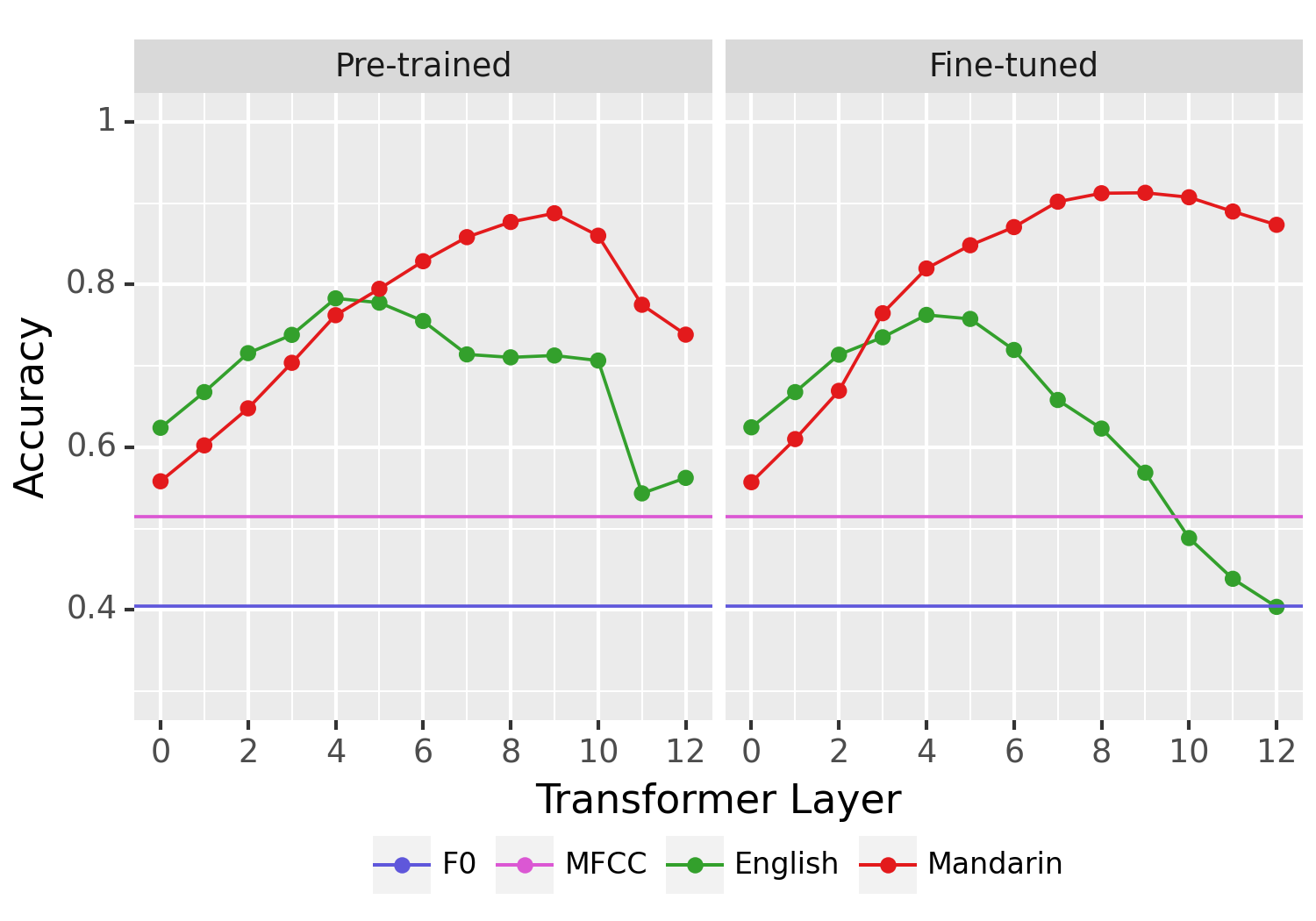}
    \caption{Classification accuracy of Mandarin lexical tones using layer-wise representations from models pre-trained and fine-tuned on Mandarin and English.}
    \label{fig:ptvsftTone}
  \includegraphics[width = \linewidth]{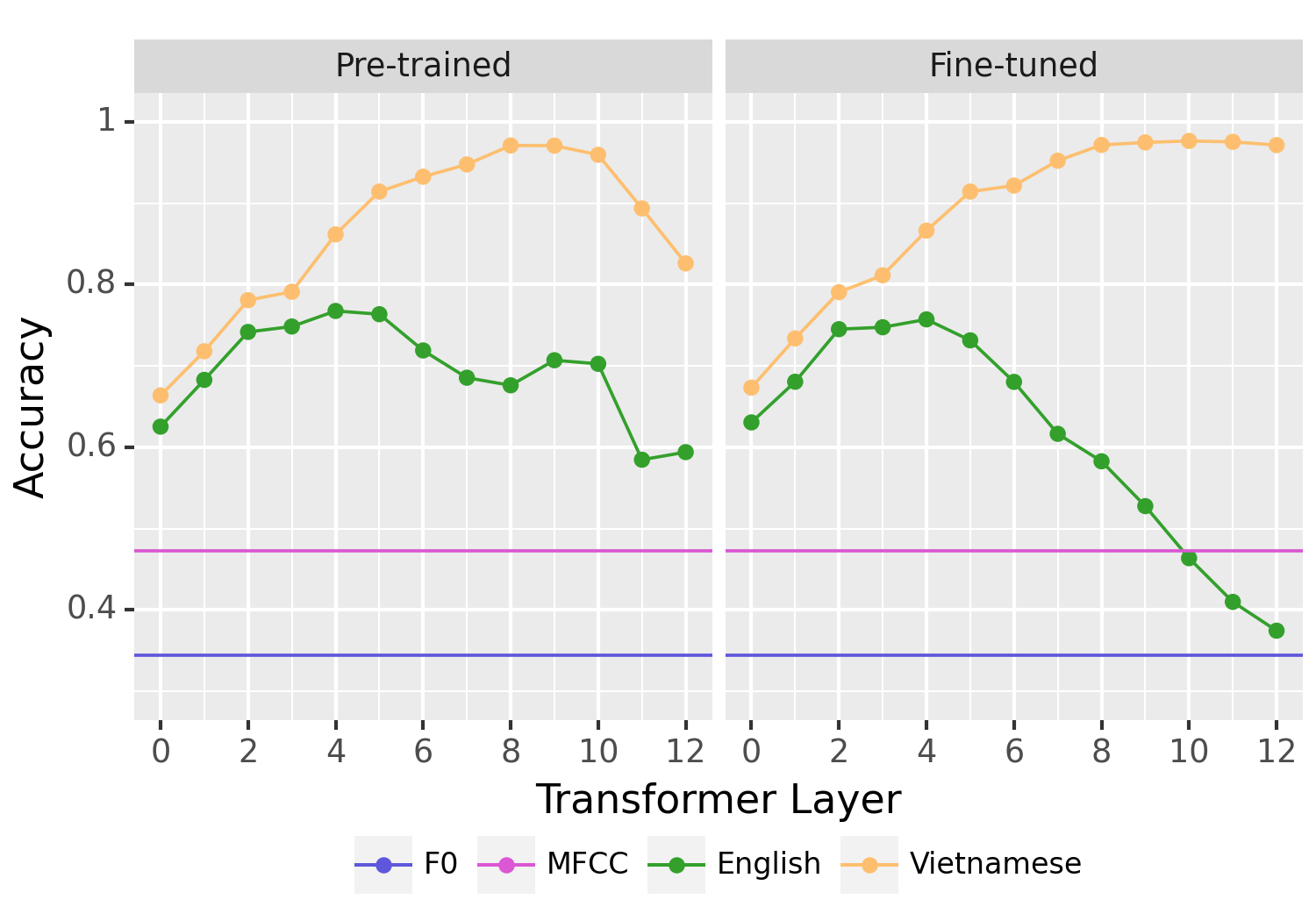}
  \caption{Classification accuracy of Vietnamese lexical tones using layer-wise representations from models pre-trained and fine-tuned on Vietnamese and English.}
  \label{fig:ptvsft-vietnamese}
\end{figure}

We examine how fine-tuning for ASR impacts the encoding of tone in SLMs. Since
tonal information is crucial for correctly transcribing tonal language input,
SLMs fine-tuned for tonal languages are expected to perform better at our tone
classification task. \Cref{fig:ptvsftTone} compares the tone classification
accuracy of the English and Mandarin models, pre-trained only (left) versus
pre-trained and then fine-tuned (right); \Cref{fig:ptvsft-vietnamese} shows the
corresponding results for English and Vietnamese.

We find that fine-tuning affects the encoding of tone for
non-tonal vs tonal language models in opposite ways: 
classification accuracy benefits from fine-tuning for Mandarin, but
is harmed by it for the English model.
The same pattern holds for English vs Vietnamese on the Vietnamese
tone data in \Cref{fig:ptvsft-vietnamese}.

These results likely reflect the fact that ASR fine-tuning encourages the SLM to increase its specialization in identifying the 
language-specific information needed to output the written form of
the language. Tonal information may not contribute much to this objective in
non-tonal languages, and thus fine-tuning would tend to remove it. In
tonal-language ASR however, tone information may be
crucial to correctly transcribe the input audio,  for example, when 
disambiguating Mandarin syllables that consist of the same segmental phonemes
and only differ in tone, in order to output the correct Chinese character.

\subsection{Comparison to human perception}
\label{sec:human}
In this section we report the results motivated by tone and consonant
perception patterns in humans.

\subsubsection{Learning trajectory}
\label{sec:trajectory}
  Children have a higher sensitivity to tone than consonant distinctions early on. For children speaking a non-tonal language, this sensitivity towards tone continues longer than sensitivity towards non-native segmental features, i.e.\ consonants and vowels
  \autocite{shiPerceptionRepresentationLexical2017,liuPerceptionTonesInfants2014}.

\begin{figure}[!htpb]
   \centering
   \includegraphics[width=0.45\textwidth]{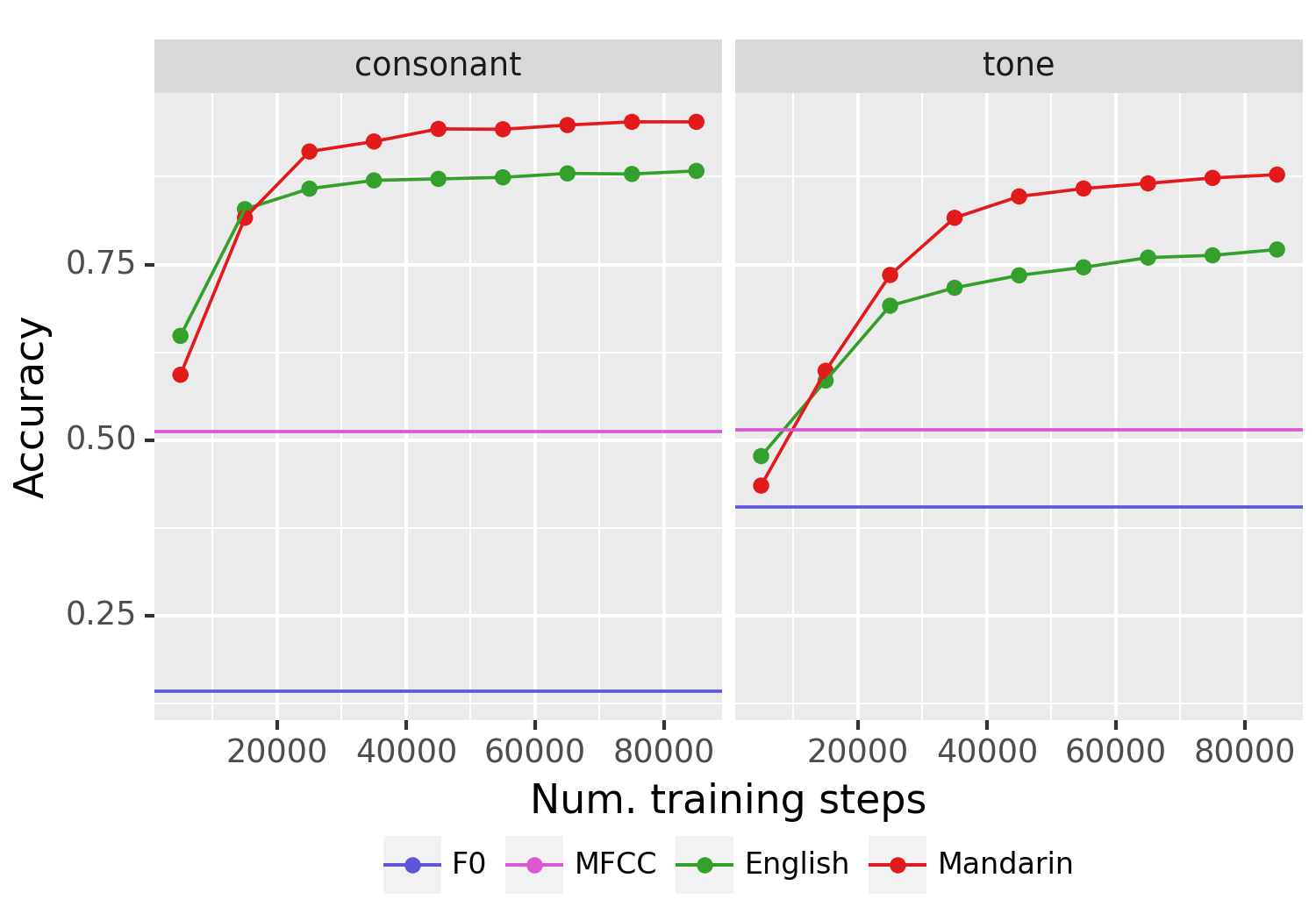}
   \caption{Classification accuracy of Mandarin lexical tones versus consonants for models pre-trained on English and Mandarin.}
   \label{fig:pretrainTvC}
 \end{figure}

Here we aim to determine the corresponding learning trajectory in
SLMs by testing them during pre-training.
\Cref{fig:pretrainTvC} shows the accuracy of classifying Mandarin consonants and
tones in the best performing layer of SLMs trained on English and Mandarin as a
function of the number of training steps. 

Although we observe classification accuracy of the SLMs quickly surpasses the F0 and
MFCC baselines after 10,000 steps, we do not detect an obvious difference in the
overall pattern between the case of consonants and tones. This suggest that SLMs
do not follow the same differential trajectory as children, at least as measured
via our methodology.

\label{sec:contrast}
\begin{figure*}[!htbp]
  \includegraphics[width = \linewidth]{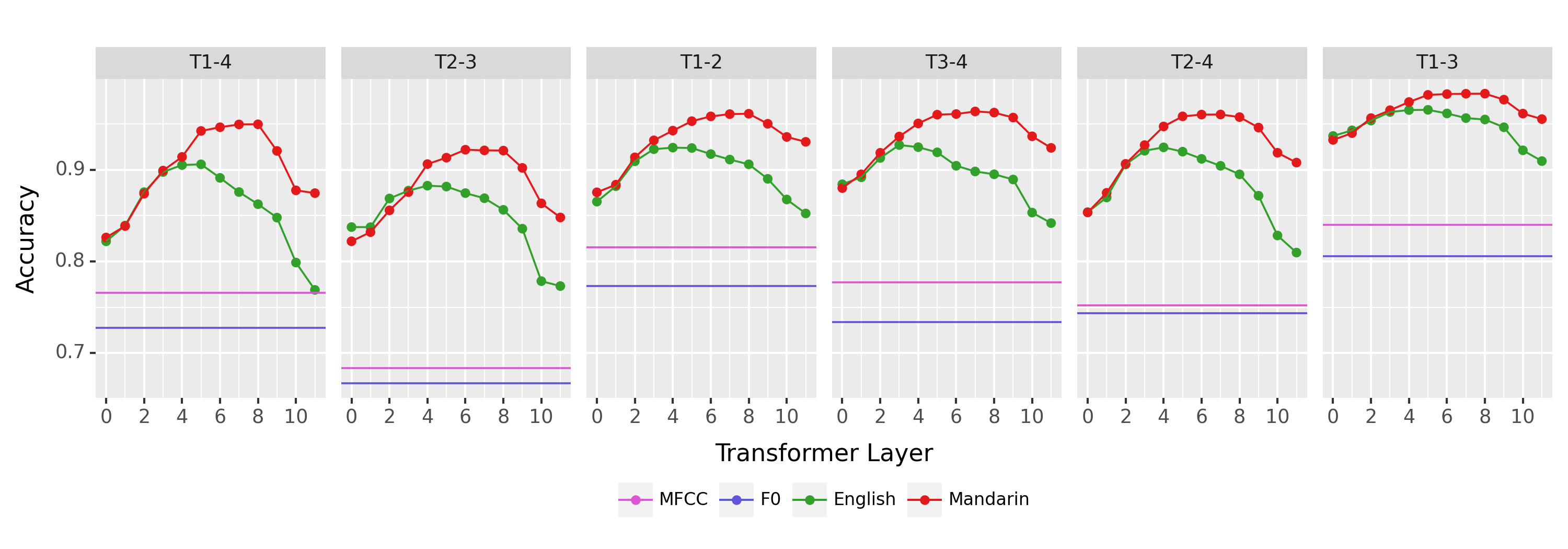}
  \caption{Binary classification accuracy for Mandarin tonal pairs,
    for English and Mandarin models.}
  \label{fig:pretrainSubclassTone}
\end{figure*}

\begin{figure}[!htbp]
  \includegraphics[width = \linewidth]{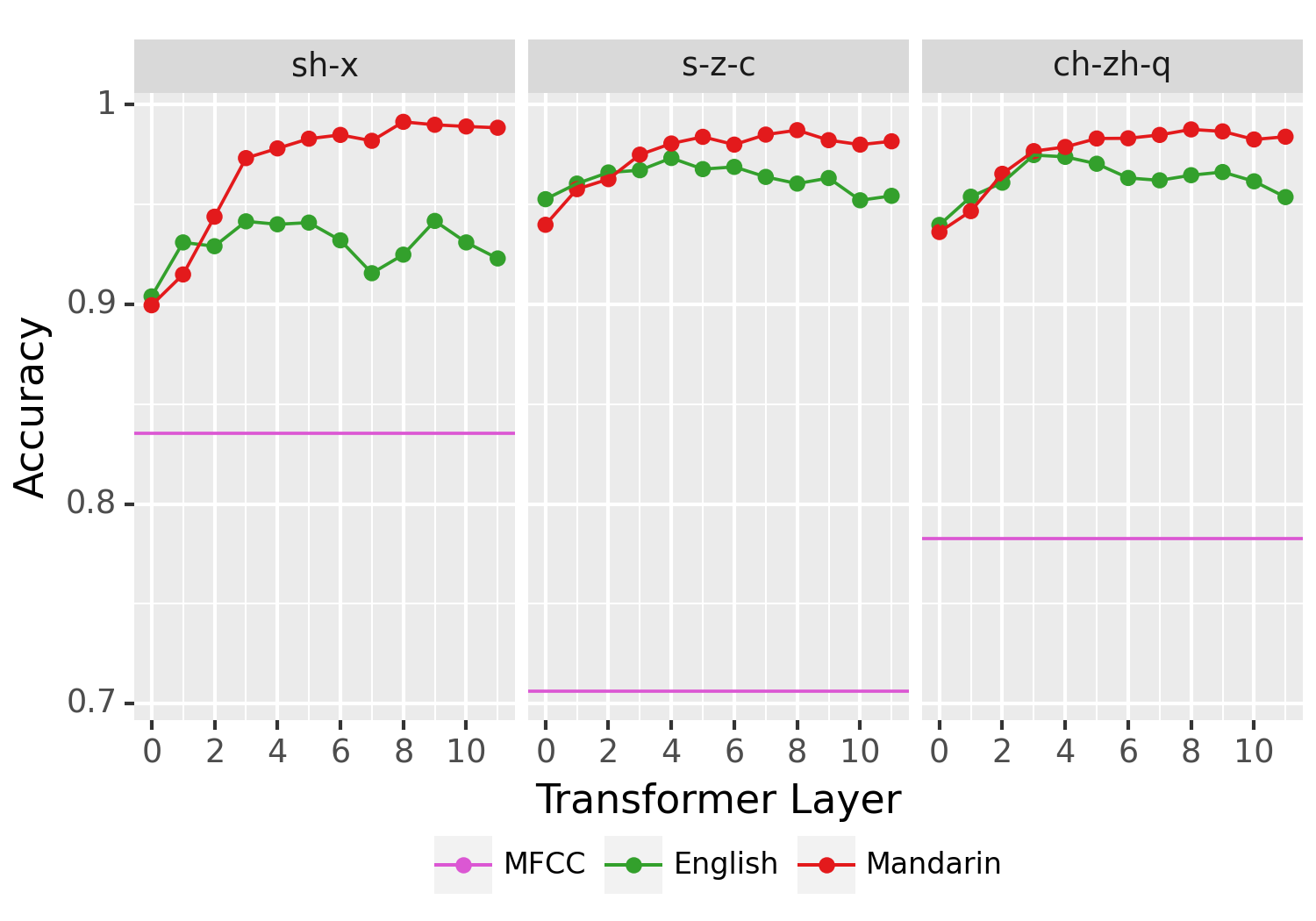}
  \caption{Classification accuracy for Mandarin consonant groups,
    for English and Mandarin models. The F0 baseline with its much lower
    classification accuracy is omitted from this figure for clarity.}
  \label{fig:pretrainSubclassConsonant}
\end{figure}

\subsubsection{Tone and consonant contrasts}
Non-native speakers can have difficulty distinguishing between T2-T3
and T1-T4 tone pairs in Mandarin
\autocite{haoSecondLanguageAcquisition2012}. While native adult listeners
have shown near perfect tone identification accuracy
\autocite{soCrosslanguagePerceptionNonnative2010,tsukadaPerceptionMandarinLexical2019},
there is literature documenting native Mandarin-learning toddlers and also some
adults having more difficulty in distinguishing the tone pair T2-3 compared to other
combinations of Mandarin lexical tone.
\textcite{huangLanguageSpecificitySpeech2011} showed that the tone pair T2-3 is the
most confusable for both native speakers of American English and Mandarin while
T1-4 also being confusable for native Mandarin speakers. We investigate this
pattern in pre-trained SLMs via a dedicated probing experiment,
using the final (85,000 steps) checkpoint of the pre-trained models
in \Cref{sec:trajectory}.
As can be seen in \Cref{fig:pretrainSubclassTone}, tone pairs T1-4 and T2-3 show the largest
differences in the best classification accuracy between the Mandarin
and English models, which roughly matches human perceptual pattern.

We complement the results on the development of tone contrast with a
parallel experiment on those Mandarin consonant contrasts which are
challenging for speakers of English. Each member of a contrasting
group is perceived as the same phoneme by English speakers due to perceptual
assimilation \autocite{wangAcquisitionMandarinConsonants2020}. 
\Cref{tab:consonantMapping} displays the resulting mapping to English
phoneme categories.

\begin{table}[htb]
  \centering
  \begin{tabular}{c|cc|cc}
  &\multicolumn{2}{c}{Mandarin} & \multicolumn{2}{c}{English} \\
  \midrule
  Group &Pinyin  & IPA & Alphabet & IPA \\
  1 & sh, x &  \textipa{\:s},  \textipa{C}  & sh & \textipa{S}\\
  2&ch, zh, q & \textipa{t\:s\super h}, \textipa{t\:s}, \textipa{tC\super h}&  ch & \textipa{tS}\\
  3& s, z, c & \textipa{s}, \textipa{ts}, \textipa{ts\super h} & s & \textipa{s}\\
\bottomrule
\end{tabular}
\caption{Perceptual mapping of Mandarin consonants onto English
  consonants \autocite{wangAcquisitionMandarinConsonants2020}.}  
\label{tab:consonantMapping}
\end{table}

\Cref{fig:pretrainSubclassConsonant} shows that accuracy for consonant
groups 2 and 3 match closely for the two models. Group 1 shows a
discrepancy, possibly due to the potential mapping of Mandarin x /\textipa{C}/ into two
English consonants sh /\textipa{S}/ and z /z/, as hypothesized by \textcite{wangAcquisitionMandarinConsonants2020}.

\section{Conclusion}
We analyze the tone encoding capabilities of spoken language models trained on
three tonal and two non-tonal languages, using classifier probes with data from
two tonal languages: Mandarin and Vietnamese. We find that SLMs trained on either
tonal or non-tonal languages encode tonal information in Mandarin and Vietnamese
to a significant degree.

We also find that fine-tuning for the speech recognition task enhances the tone
encoding capabilities of models trained on tonal languages but reduces them for
models trained on non-tonal languages. While we see evidence suggesting that the
learning trajectories of SLMs in pre-training do not follow the same
developmental trajectories found in human language acquisition, we find that
SLMs show patterns similar to that of human listeners in tone and consonant
perception experiments. 

While this paper focused on investigating the encoding of lexical tones with
Mandarin and Vietnamese as case studies, Mandarin and Vietnamese are in no way
representative of all the diverse tonal languages that exist in the world. The
encoding of other suprasegmental features such as stress patterns and intonation
is equally important to study in future work. We hope that explaining
self-supervised models of spoken languages provides a unique perspective for us
to contribute to a better understanding of how languages work. Given the rise of
SLMs trained with multilingual datasets
\autocite{conneauUnsupervisedCrossLingualRepresentation2021,radfordRobustSpeechRecognition2022},
it would be interesting to investigate if the multilingual SLMs encode
segmentals and suprasegmentals from different languages more robustly. This
paper serves as a starting point in the research of the encoding of
suprasegmental cues in spoken language models. 

\section{Limitations}
We selected SLMs based on the wav2vec2 architecture in our experimental design,
but we acknowledge that the training data of the models selected is quite varied
in their size and quality (noisy vs clean speech) as described in
\Cref{sec:trainingData}. This is partially due to the scarce availability of
(high-quality) speech data for underrepresented languages, especially the many
tonal languages of the world. Hence SLMs pre-trained on monolingual datasets of
these languages are also sparse. The Vietnamese wav2vec2 model
\autocite{Thai_Binh_Nguyen_wav2vec2_vi_2021} was trained on a significantly
larger amount of data (13k hours) than the other models tested (around 1000
hours). It is possible that in addition to the inclusion of tonal languages in
training, the amount of training data also played a role in increasing the tonal
encoding capabilities of SLMs. However, literature has shown that more training
data does not always have a positive impact on the models performance if the
additional data is noisy
\autocite{parcolletLeBenchmarkStandardizedReplicable2023}.
At the same time, we note
that the Cantonese model \autocite{huangWav2vecASRCantoneseSpeaking2023}, in
addition to being pre-trained on a larger dataset, is also different in architecture. 
Additionally, our use of two read speech datasets as test data does not fully
reflect the linguistic diversity of different accents and dialects in Mandarin and
Vietnamese. Future work needs to go wider and deeper in both model architecture
and dataset diversity in order to uncover more generalizable patterns in
different languages.

\section{Acknowledgements}
This publication is part of the project \textit{InDeep: Interpreting Deep
Learning Models for Text and Sound} (with project number NWA.1292.19.399) of the
National Research Agenda (NWA-ORC) program. Funding by the Dutch Research
Council (NWO) is gratefully acknowledged. We also thank SURF (www.surf.nl) for
the support in using the National Supercomputer Snellius.

\bibliography{UVT}

\begin{thebibliography}{69}
\expandafter\ifx\csname natexlab\endcsname\relax\def\natexlab#1{#1}\fi

\bibitem[{Abdullah et~al.(2021)Abdullah, Zaitova, Avgustinova, M{\"o}bius, and
  Klakow}]{abdullahHowFamiliarDoes2021}
Badr~M. Abdullah, Iuliia Zaitova, Tania Avgustinova, Bernd M{\"o}bius, and
  Dietrich Klakow. 2021.
\newblock \href {https://doi.org/10.48550/arXiv.2109.10179} {How {{Familiar
  Does That Sound}}? {{Cross-Lingual Representational Similarity Analysis}} of
  {{Acoustic Word Embeddings}}}.

\bibitem[{Baevski et~al.(2020)Baevski, Zhou, Mohamed, and
  Auli}]{baevskiWav2vecFrameworkSelfSupervised2020}
Alexei Baevski, Henry Zhou, Abdelrahman Mohamed, and Michael Auli. 2020.
\newblock \href {https://doi.org/10.48550/ARXIV.2006.11477} {Wav2vec 2.0: {{A
  Framework}} for {{Self-Supervised Learning}} of {{Speech Representations}}}.

\bibitem[{Bartelds et~al.(2022)Bartelds, {de Vries}, Sanal, Richter, Liberman,
  and Wieling}]{barteldsNeuralRepresentationsModeling2022}
Martijn Bartelds, Wietse {de Vries}, Faraz Sanal, Caitlin Richter, Mark
  Liberman, and Martijn Wieling. 2022.
\newblock \href {https://doi.org/10.1016/j.wocn.2022.101137} {Neural
  representations for modeling variation in speech}.
\newblock \emph{Journal of Phonetics}, 92:101137.

\bibitem[{{Belotel-Grenie} and
  Grenie(1994)}]{Belotel-Grenie1994PHONATIONCHINESE}
Agnes {Belotel-Grenie} and Michel Grenie. 1994.
\newblock \href {https://doi.org/10.21437/icslp.1994-89} {Phonation types
  analysis in standard chinese}.
\newblock In \emph{3rd International Conference on Spoken Language Processing,
  {{ICSLP}} 1994}, pages 343--346. {The International Society for Computers and
  Their Applications (ISCA)}.

\bibitem[{Boersma and Weenink(2021)}]{boersmaPraatDoingPhonetics2021}
Paul Boersma and David Weenink. 2021.
\newblock Praat: Doing phonetics by computer [{{Computer}} program].

\bibitem[{Brunelle(2009)}]{Brunelle2009ToneVietnamese}
Marc Brunelle. 2009.
\newblock \href {https://doi.org/10.1016/j.wocn.2008.09.003} {Tone perception
  in northern and southern vietnamese}.
\newblock \emph{Journal of Phonetics}, 37(1):79--96.

\bibitem[{Bu et~al.(2017)Bu, Du, Na, Wu, and
  Zheng}]{buAISHELL1OpenSourceMandarin2017a}
Hui Bu, Jiayu Du, Xingyu Na, Bengu Wu, and Hao Zheng. 2017.
\newblock \href {https://doi.org/10.48550/arXiv.1709.05522} {{{AISHELL-1}}:
  {{An Open-Source Mandarin Speech Corpus}} and {{A Speech Recognition
  Baseline}}}.

\bibitem[{Chai(2019)}]{chaiSOURCECREAKMANDARIN2019}
Yuan Chai. 2019.
\newblock {{The source of creak in Mandarin utterances}}.
\newblock In \emph{Proceedings of ICPhS 2019}.

\bibitem[{Chen(2000)}]{chenToneSandhiPatterns2000}
Matthew~Y. Chen. 2000.
\newblock \href {https://doi.org/10.1017/CBO9780511486364} {\emph{Tone
  {{Sandhi}}: {{Patterns}} across {{Chinese Dialects}}}}.
\newblock Cambridge {{Studies}} in {{Linguistics}}. {Cambridge University
  Press}, {Cambridge}.

\bibitem[{Chen et~al.(2022)Chen, Gao, and
  Xu}]{chenComputationalModellingTone2022}
Yue Chen, Yingming Gao, and Yi~Xu. 2022.
\newblock \href {https://doi.org/10.3390/brainsci12030337} {Computational
  {{Modelling}} of {{Tone Perception Based}} on {{Direct Processing}} of f0
  {{Contours}}}.
\newblock \emph{Brain Sciences}, 12(3):337.

\bibitem[{Conneau et~al.(2021)Conneau, Baevski, Collobert, Mohamed, and
  Auli}]{conneauUnsupervisedCrossLingualRepresentation2021}
Alexis Conneau, Alexei Baevski, Ronan Collobert, Abdelrahman Mohamed, and
  Michael Auli. 2021.
\newblock \href {https://doi.org/10.21437/Interspeech.2021-329} {Unsupervised
  {{Cross-Lingual Representation Learning}} for {{Speech Recognition}}}.
\newblock In \emph{Interspeech 2021}, pages 2426--2430. {ISCA}.

\bibitem[{Conneau et~al.(2018)Conneau, Kruszewski, Lample, Barrault, and
  Baroni}]{conneauWhatYouCan2018}
Alexis Conneau, German Kruszewski, Guillaume Lample, Lo{\"i}c Barrault, and
  Marco Baroni. 2018.
\newblock \href {https://doi.org/10.18653/v1/P18-1198} {What you can cram into
  a single \$\&!\#* vector: {{Probing}} sentence embeddings for linguistic
  properties}.
\newblock In \emph{Proceedings of the 56th {{Annual Meeting}} of the
  {{Association}} for {{Computational Linguistics}} ({{Volume}} 1: {{Long
  Papers}})}, pages 2126--2136, {Melbourne, Australia}. {Association for
  Computational Linguistics}.

\bibitem[{Cruz~Bland{\'o}n et~al.(2023)Cruz~Bland{\'o}n, Cristia, and
  R{\"a}s{\"a}nen}]{cruzblandonIntroducingMetaanalysisEvaluation2023}
Mar{\'i}a~Andrea Cruz~Bland{\'o}n, Alejandrina Cristia, and Okko
  R{\"a}s{\"a}nen. 2023.
\newblock \href {https://doi.org/10.1111/cogs.13307} {Introducing
  {{Meta-analysis}} in the {{Evaluation}} of {{Computational Models}} of
  {{Infant Language Development}}}.
\newblock \emph{Cognitive Science}, 47(7):e13307.

\bibitem[{{de Seyssel} et~al.(2022){de Seyssel}, Lavechin, Adi, Dupoux, and
  Wisniewski}]{deSeyssel2022ProbingPL}
Maureen {de Seyssel}, Marvin Lavechin, Yossi Adi, Emmanuel Dupoux, and
  Guillaume Wisniewski. 2022.
\newblock Probing phoneme, language and speaker information in unsupervised
  speech representations.
\newblock In \emph{Interspeech}.

\bibitem[{Devlin et~al.(2019)Devlin, Chang, Lee, and
  Toutanova}]{devlinBERTPretrainingDeep2019}
Jacob Devlin, Ming-Wei Chang, Kenton Lee, and Kristina Toutanova. 2019.
\newblock \href {https://doi.org/10.48550/arXiv.1810.04805} {{{BERT}}:
  {{Pre-training}} of {{Deep Bidirectional Transformers}} for {{Language
  Understanding}}}.

\bibitem[{DiCanio(2012)}]{dicanioCrosslinguisticPerceptionItunyoso2012}
Christian~T. DiCanio. 2012.
\newblock \href {https://doi.org/10.1016/j.wocn.2012.05.003} {Cross-linguistic
  perception of {{Itunyoso Trique}} tone}.
\newblock \emph{Journal of Phonetics}, 40(5):672--688.

\bibitem[{Du et~al.(2018)Du, Na, Liu, and
  Bu}]{duAISHELL2TransformingMandarin2018}
Jiayu Du, Xingyu Na, Xuechen Liu, and Hui Bu. 2018.
\newblock \href {https://doi.org/10.48550/arXiv.1808.10583} {{{AISHELL-2}}:
  {{Transforming Mandarin ASR Research Into Industrial Scale}}}.

\bibitem[{Hao(2012)}]{haoSecondLanguageAcquisition2012}
Yen-Chen Hao. 2012.
\newblock \href {https://doi.org/10.1016/j.wocn.2011.11.001} {Second language
  acquisition of {{Mandarin Chinese}} tones by tonal and non-tonal language
  speakers}.
\newblock \emph{Journal of Phonetics}, 40(2):269--279.

\bibitem[{Hewitt and Manning(2019)}]{hewittStructuralProbeFinding2019}
John Hewitt and Christopher~D. Manning. 2019.
\newblock \href {https://doi.org/10.18653/v1/N19-1419} {A {{Structural Probe}}
  for {{Finding Syntax}} in {{Word Representations}}}.
\newblock In \emph{Proceedings of the 2019 {{Conference}} of the {{North
  American Chapter}} of the {{Association}} for {{Computational Linguistics}}:
  {{Human Language Technologies}}, {{Volume}} 1 ({{Long}} and {{Short
  Papers}})}, pages 4129--4138, {Minneapolis, Minnesota}. {Association for
  Computational Linguistics}.

\bibitem[{Hsu et~al.(2021)Hsu, Bolte, Tsai, Lakhotia, Salakhutdinov, and
  Mohamed}]{hsuHuBERTSelfSupervisedSpeech2021}
Wei-Ning Hsu, Benjamin Bolte, Yao-Hung~Hubert Tsai, Kushal Lakhotia, Ruslan
  Salakhutdinov, and Abdelrahman Mohamed. 2021.
\newblock \href {http://arxiv.org/abs/2106.07447} {{{HuBERT}}:
  {{Self-Supervised Speech Representation Learning}} by {{Masked Prediction}}
  of {{Hidden Units}}}.

\bibitem[{Huang and Mak(2023)}]{huangWav2vecASRCantoneseSpeaking2023}
Ranzo Huang and Brian Mak. 2023.
\newblock \href {https://doi.org/10.21437/Interspeech.2023-2470} {Wav2vec 2.0
  {{ASR}} for {{Cantonese-Speaking Older Adults}} in a {{Clinical Setting}}}.
\newblock In \emph{{{INTERSPEECH}} 2023}, pages 4958--4962. {ISCA}.

\bibitem[{Huang and Johnson(2011)}]{huangLanguageSpecificitySpeech2011}
Tsan Huang and Keith Johnson. 2011.
\newblock \href {https://doi.org/10.1159/000327392} {Language {{Specificity}}
  in {{Speech Perception}}: {{Perception}} of {{Mandarin Tones}} by {{Native}}
  and {{Nonnative Listeners}}}.
\newblock \emph{Phonetica}, 67(4):243--267.

\bibitem[{Huang(2020)}]{huangDifferentAttributesCreaky2020}
Yaqian Huang. 2020.
\newblock \href {https://doi.org/10.1121/10.0000721} {Different attributes of
  creaky voice distinctly affect {{Mandarin}} tonal perception}.
\newblock \emph{The Journal of the Acoustical Society of America},
  147(3):1441--1458.

\bibitem[{Hyman(2018)}]{hymanWhatToneTeaches2018}
Larry~M. Hyman. 2018.
\newblock What tone teaches us about language.
\newblock \emph{Language}, 94(3):698--709.

\bibitem[{Jadoul et~al.(2018)Jadoul, Thompson, and {de
  Boer}}]{jadoulIntroducingParselmouthPython2018}
Yannick Jadoul, Bill Thompson, and Bart {de Boer}. 2018.
\newblock \href {https://doi.org/10.1016/j.wocn.2018.07.001} {Introducing
  {{Parselmouth}}: {{A Python}} interface to {{Praat}}}.
\newblock \emph{Journal of Phonetics}, 71:1--15.

\bibitem[{Jun and Kubozono(2020)}]{junAsianPacificRim2020}
Sun-Ah Jun and Haruo Kubozono. 2020.
\newblock Asian {{Pacific Rim}}.
\newblock In Carlos Gussenhoven and Aoju Chen, editors, \emph{The {{Oxford
  Handbook}} of {{Language Prosody}}}, Oxford {{Handbooks}}. {Oxford University
  Press}, {Oxford, New York}.

\bibitem[{Kirby(2008)}]{kirbyVPhonVietnamesePhonetizer2008}
James Kirby. 2008.
\newblock {{vPhon}}: A {{Vietnamese}} phonetizer (version 2.1.1).

\bibitem[{Kirby(2011)}]{kirbyVietnameseHanoiVietnamese2011}
James~P. Kirby. 2011.
\newblock \href {https://doi.org/10.1017/S0025100311000181} {Vietnamese (hanoi
  vietnamese)}.
\newblock \emph{Journal of the international phonetic association},
  41(3):381--392.

\bibitem[{Kuang(2017)}]{kuangCovariationVoiceQuality2017}
Jianjing Kuang. 2017.
\newblock \href {https://doi.org/10.1121/1.5003649} {Covariation between voice
  quality and pitch: {{Revisiting}} the case of {{Mandarin}} creaky voice}.
\newblock \emph{The Journal of the Acoustical Society of America},
  142(3):1693--1706.

\bibitem[{Lavechin et~al.(2023)Lavechin, De~Seyssel, M{\'e}tais, Metze,
  Mohamed, Bredin, Dupoux, and Cristia}]{lavechinStatisticalLearningModels2023}
Marvin Lavechin, Maureen De~Seyssel, Marianne M{\'e}tais, Florian Metze,
  Abdelrahman Mohamed, Herv{\'e} Bredin, Emmanuel Dupoux, and Alejandrina
  Cristia. 2023.
\newblock Statistical learning models of early phonetic acquisition struggle
  with child-centered audio data.

\bibitem[{Liu and Kager(2014)}]{liuPerceptionTonesInfants2014}
Liquan Liu and Ren{\'e} Kager. 2014.
\newblock \href {https://doi.org/10.1016/j.cognition.2014.06.004} {Perception
  of tones by infants learning a non-tone language}.
\newblock \emph{Cognition}, 133(2):385--394.

\bibitem[{Lu and Chen(2022)}]{luContextawareKnowledgeTransferring2022}
Ke-Han Lu and Kuan-Yu Chen. 2022.
\newblock \href {https://doi.org/10.48550/arXiv.2210.06244} {A context-aware
  knowledge transferring strategy for {{CTC-based ASR}}}.

\bibitem[{Luong and Vu(2016)}]{luong-vu-2016-non}
Hieu-Thi Luong and Hai-Quan Vu. 2016.
\newblock A non-expert {{Kaldi}} recipe for {{Vietnamese}} speech recognition
  system.
\newblock In \emph{Proceedings of the Third International Workshop on Worldwide
  Language Service Infrastructure and Second Workshop on Open Infrastructures
  and Analysis Frameworks for Human Language Technologies
  ({{WLSI}}/{{OIAF4HLT2016}})}, pages 51--55, {Osaka, Japan}. {The COLING 2016
  Organizing Committee}.

\bibitem[{Ma et~al.(2021)Ma, Ryant, and
  Liberman}]{maProbingAcousticRepresentations2021}
Danni Ma, Neville Ryant, and Mark Liberman. 2021.
\newblock \href {https://doi.org/10.1109/ICASSP39728.2021.9414776} {Probing
  {{Acoustic Representations}} for {{Phonetic Properties}}}.
\newblock In \emph{{{ICASSP}} 2021 - 2021 {{IEEE International Conference}} on
  {{Acoustics}}, {{Speech}} and {{Signal Processing}} ({{ICASSP}})}, pages
  311--315.

\bibitem[{Magic Data Technology~Co.(2019)}]{magicdata2019}
{\relax Ltd}.~Magic Data Technology~Co. 2019.
\newblock {{MAGICDATA Mandarin Chinese Read Speech Corpus}}.

\bibitem[{Martin et~al.(2023)Martin, Gauthier, Breiss, and
  Levy}]{martinProbingSelfsupervisedSpeech2023}
Kinan Martin, Jon Gauthier, Canaan Breiss, and Roger Levy. 2023.
\newblock \href {https://doi.org/10.21437/Interspeech.2023-2359} {Probing
  {{Self-supervised Speech Models}} for {{Phonetic}} and {{Phonemic
  Information}}: {{A Case Study}} in {{Aspiration}}}.
\newblock In \emph{{{INTERSPEECH}} 2023}, pages 251--255. {ISCA}.

\bibitem[{McAuliffe et~al.(2017)McAuliffe, Socolof, Mihuc, Wagner, and
  Sonderegger}]{mcauliffeMontrealForcedAligner2017}
Michael McAuliffe, Michaela Socolof, Sarah Mihuc, Michael Wagner, and Morgan
  Sonderegger. 2017.
\newblock \href {https://doi.org/10.21437/Interspeech.2017-1386} {Montreal
  {{Forced Aligner}}: {{Trainable Text-Speech Alignment Using Kaldi}}}.
\newblock In \emph{Interspeech 2017}, pages 498--502. {ISCA}.

\bibitem[{McFee et~al.(2023)McFee, McVicar, Faronbi, Roman, Gover, Balke,
  Seyfarth, Malek, Raffel, Lostanlen, Van~Niekirk, Lee, Cwitkowitz, Zalkow,
  Nieto, Ellis, Mason, {Kyungyun Lee}, Steers, Halvachs, Thom{\'e},
  {Robert-St{\"o}ter}, Bittner, {Ziyao Wei}, Weiss, Battenberg, {Keunwoo Choi},
  Yamamoto, {CJ Carr}, Metsai, Sullivan, Friesch, {Asmitha Krishnakumar},
  Hidaka, Kowalik, Keller, Mazur, {Chabot-Leclerc}, Hawthorne, {Chandrashekhar
  Ramaprasad}, {Myungchul Keum}, Gomez, Monroe, Morozov, Eliasi,
  {Nullmightybofo}, Biberstein, {N. Dorukhan Sergin}, Hennequin, Naktinis,
  {Beantowel}, Kim, {\AA}sen, Lim, Malins, Here{\~n}{\'u}, Van Der~Struijk,
  Nickel, Wu, Wang, Gates, Vollrath, Sarroff, {Xiao-Ming}, Porter, Kranzler,
  {Voodoohop}, Di~Gangi, Jinoz, Guerrero, {Abduttayyeb Mazhar}, {Toddrme2178},
  Baratz, Kostin, {Xinlu Zhuang}, {Cash TingHin Lo}, Campr, Semeniuc, Biswal,
  {Shayenne Moura}, Brossier, {Hojin Lee}, and
  Pimenta}]{mcfeeLibrosaLibrosa102023}
Brian McFee, Matt McVicar, Daniel Faronbi, Iran Roman, Matan Gover, Stefan
  Balke, Scott Seyfarth, Ayoub Malek, Colin Raffel, Vincent Lostanlen, Benjamin
  Van~Niekirk, Dana Lee, Frank Cwitkowitz, Frank Zalkow, Oriol Nieto, Dan
  Ellis, Jack Mason, {Kyungyun Lee}, Bea Steers, Emily Halvachs, Carl
  Thom{\'e}, Fabian {Robert-St{\"o}ter}, Rachel Bittner, {Ziyao Wei}, Adam
  Weiss, Eric Battenberg, {Keunwoo Choi}, Ryuichi Yamamoto, {CJ Carr}, Alex
  Metsai, Stefan Sullivan, Pius Friesch, {Asmitha Krishnakumar}, Shunsuke
  Hidaka, Steve Kowalik, Fabian Keller, Dan Mazur, Alexandre {Chabot-Leclerc},
  Curtis Hawthorne, {Chandrashekhar Ramaprasad}, {Myungchul Keum}, Juanita
  Gomez, Will Monroe, Viktor~Andreevitch Morozov, Kian Eliasi,
  {Nullmightybofo}, Paul Biberstein, {N. Dorukhan Sergin}, Romain Hennequin,
  Rimvydas Naktinis, {Beantowel}, Taewoon Kim, Jon~Petter {\AA}sen, Joon Lim,
  Alex Malins, Dar{\'i}o Here{\~n}{\'u}, Stef Van Der~Struijk, Lorenz Nickel,
  Jackie Wu, Zhen Wang, Tim Gates, Matt Vollrath, Andy Sarroff, {Xiao-Ming},
  Alastair Porter, Seth Kranzler, {Voodoohop}, Mattia Di~Gangi, Helmi Jinoz,
  Connor Guerrero, {Abduttayyeb Mazhar}, {Toddrme2178}, Zvi Baratz, Anton
  Kostin, {Xinlu Zhuang}, {Cash TingHin Lo}, Pavel Campr, Eric Semeniuc, Monsij
  Biswal, {Shayenne Moura}, Paul Brossier, {Hojin Lee}, and Waldir Pimenta.
  2023.
\newblock \href {https://doi.org/10.5281/ZENODO.8252662} {Librosa/librosa:
  0.10.1}.
\newblock Zenodo.

\bibitem[{Mehler et~al.(1988)Mehler, Jusczyk, Lambertz, Halsted, Bertoncini,
  and {Amiel-Tison}}]{mehlerPrecursorLanguageAcquisition1988}
J.~Mehler, P.~Jusczyk, G.~Lambertz, N.~Halsted, J.~Bertoncini, and
  C.~{Amiel-Tison}. 1988.
\newblock \href {https://doi.org/10.1016/0010-0277(88)90035-2} {A precursor of
  language acquisition in young infants}.
\newblock \emph{Cognition}, 29(2):143--178.

\bibitem[{Nazzi et~al.(1998)Nazzi, Bertoncini, and
  Mehler}]{nazziLanguageDiscriminationNewborns1998}
T.~Nazzi, J.~Bertoncini, and J.~Mehler. 1998.
\newblock \href {https://doi.org/10.1037//0096-1523.24.3.756} {Language
  discrimination by newborns: Toward an understanding of the role of rhythm}.
\newblock \emph{Journal of Experimental Psychology. Human Perception and
  Performance}, 24(3):756--766.

\bibitem[{Nguyen(2021)}]{Thai_Binh_Nguyen_wav2vec2_vi_2021}
Thai~Binh Nguyen. 2021.
\newblock \href {https://doi.org/10.5281/zenodo.5356039} {Vietnamese end-to-end
  speech recognition using wav2vec 2.0}.

\bibitem[{Ott et~al.(2019)Ott, Edunov, Baevski, Fan, Gross, Ng, Grangier, and
  Auli}]{ottFairseqFastExtensible2019}
Myle Ott, Sergey Edunov, Alexei Baevski, Angela Fan, Sam Gross, Nathan Ng,
  David Grangier, and Michael Auli. 2019.
\newblock \href {https://doi.org/10.18653/v1/N19-4009} {Fairseq: {{A Fast}},
  {{Extensible Toolkit}} for {{Sequence Modeling}}}.
\newblock pages 48--53, {Minneapolis, Minnesota}. {Association for
  Computational Linguistics}.

\bibitem[{Panayotov et~al.(2015)Panayotov, Chen, Povey, and
  Khudanpur}]{panayotovLibrispeechASRCorpus2015}
Vassil Panayotov, Guoguo Chen, Daniel Povey, and Sanjeev Khudanpur. 2015.
\newblock \href {https://doi.org/10.1109/ICASSP.2015.7178964} {Librispeech:
  {{An ASR}} corpus based on public domain audio books}.
\newblock In \emph{2015 {{IEEE International Conference}} on {{Acoustics}},
  {{Speech}} and {{Signal Processing}} ({{ICASSP}})}, pages 5206--5210.

\bibitem[{Parcollet et~al.(2023)Parcollet, Nguyen, Evain, Boito, Pupier,
  Mdhaffar, Le, Alisamir, Tomashenko, Dinarelli, Zhang, Allauzen, Coavoux,
  Esteve, Rouvier, Goulian, Lecouteux, Portet, Rossato, Ringeval, Schwab, and
  Besacier}]{parcolletLeBenchmarkStandardizedReplicable2023}
Titouan Parcollet, Ha~Nguyen, Solene Evain, Marcely~Zanon Boito, Adrien Pupier,
  Salima Mdhaffar, Hang Le, Sina Alisamir, Natalia Tomashenko, Marco Dinarelli,
  Shucong Zhang, Alexandre Allauzen, Maximin Coavoux, Yannick Esteve, Mickael
  Rouvier, Jerome Goulian, Benjamin Lecouteux, Francois Portet, Solange
  Rossato, Fabien Ringeval, Didier Schwab, and Laurent Besacier. 2023.
\newblock \href {http://arxiv.org/abs/2309.05472} {{{LeBenchmark}} 2.0: A
  {{Standardized}}, {{Replicable}} and {{Enhanced Framework}} for
  {{Self-supervised Representations}} of {{French Speech}}}.

\bibitem[{Pasad et~al.(2024)Pasad, Chien, Settle, and
  Livescu}]{pasadWhatSelfSupervisedSpeech2024}
Ankita Pasad, Chung-Ming Chien, Shane Settle, and Karen Livescu. 2024.
\newblock \href {https://doi.org/10.48550/arXiv.2307.00162} {What {{Do
  Self-Supervised Speech Models Know About Words}}?}

\bibitem[{Pasad et~al.(2021)Pasad, Chou, and
  Livescu}]{pasadLayerwiseAnalysisSelfsupervised2021}
Ankita Pasad, Ju-Chieh Chou, and Karen Livescu. 2021.
\newblock \href {http://arxiv.org/abs/2107.04734} {Layer-wise {{Analysis}} of a
  {{Self-supervised Speech Representation Model}}}.

\bibitem[{Pratap et~al.(2020)Pratap, Xu, Sriram, Synnaeve, and
  Collobert}]{pratapMLSLargeScaleMultilingual2020a}
Vineel Pratap, Qiantong Xu, Anuroop Sriram, Gabriel Synnaeve, and Ronan
  Collobert. 2020.
\newblock \href {https://doi.org/10.21437/Interspeech.2020-2826} {{{MLS}}: {{A
  Large-Scale Multilingual Dataset}} for {{Speech Research}}}.
\newblock In \emph{Interspeech 2020}, pages 2757--2761.

\bibitem[{Radford et~al.(2022)Radford, Kim, Xu, Brockman, McLeavey, and
  Sutskever}]{radfordRobustSpeechRecognition2022}
Alec Radford, Jong~Wook Kim, Tao Xu, Greg Brockman, Christine McLeavey, and
  Ilya Sutskever. 2022.
\newblock Robust {{Speech Recognition}} via {{Large-Scale Weak Supervision}}.

\bibitem[{Rhee et~al.(2021)Rhee, Chen, and Kuang}]{rheeGoingF0Acquisition2021}
Nari Rhee, Aoju Chen, and Jianjing Kuang. 2021.
\newblock \href {https://doi.org/10.1017/S0305000920000239} {Going beyond
  {{F0}}: {{The}} acquisition of {{Mandarin}} tones}.
\newblock \emph{Journal of Child Language}, 48(2):387--398.

\bibitem[{Ryant et~al.(2014{\natexlab{a}})Ryant, Slaney, Liberman, Shriberg,
  and Yuan}]{ryantHighlyAccurateMandarin2014}
Neville Ryant, Malcolm Slaney, Mark Liberman, Elizabeth Shriberg, and Jiahong
  Yuan. 2014{\natexlab{a}}.
\newblock \href {https://doi.org/10.21437/SpeechProsody.2014-123} {Highly
  {{Accurate Mandarin Tone Classification In The Absence}} of {{Pitch
  Information}}}.
\newblock In \emph{Speech {{Prosody}} 2014}, pages 673--677. {ISCA}.

\bibitem[{Ryant et~al.(2014{\natexlab{b}})Ryant, Yuan, and
  Liberman}]{ryantMandarinToneClassification2014}
Neville Ryant, Jiahong Yuan, and Mark Liberman. 2014{\natexlab{b}}.
\newblock \href {https://doi.org/10.1109/ICASSP.2014.6854527} {Mandarin tone
  classification without pitch tracking}.
\newblock In \emph{2014 {{IEEE International Conference}} on {{Acoustics}},
  {{Speech}} and {{Signal Processing}} ({{ICASSP}})}, pages 4868--4872,
  {Florence, Italy}. {IEEE}.

\bibitem[{Schaefer and Darcy(2014)}]{schaeferLexicalFunctionPitch2014}
Vance Schaefer and Isabelle Darcy. 2014.
\newblock \href {https://doi.org/10.1515/lp-2014-0016} {Lexical function of
  pitch in the first language shapes cross-linguistic perception of {{Thai}}
  tones}.
\newblock \emph{Laboratory Phonology}, 5(4):489--522.

\bibitem[{Shen et~al.(2023)Shen, Alishahi, Bisazza, and
  Chrupa{\l}a}]{shenWaveSyntaxProbing2023}
Gaofei Shen, Afra Alishahi, Arianna Bisazza, and Grzegorz Chrupa{\l}a. 2023.
\newblock \href {https://doi.org/10.21437/Interspeech.2023-679} {Wave to
  {{Syntax}}: {{Probing}} spoken language models for syntax}.
\newblock In \emph{{{INTERSPEECH}} 2023}, pages 1259--1263.

\bibitem[{Shi et~al.(2017)Shi, Gao, Achim, and
  Li}]{shiPerceptionRepresentationLexical2017}
Rushen Shi, Jun Gao, Andr{\'e} Achim, and Aijun Li. 2017.
\newblock Perception and {{Representation}} of {{Lexical Tones}} in {{Native
  Mandarin-Learning Infants}} and {{Toddlers}}.
\newblock \emph{Frontiers in Psychology}, 8.

\bibitem[{Singh and Fu(2016)}]{singhNewViewLanguage2016}
Leher Singh and Charlene S.~L. Fu. 2016.
\newblock \href {https://doi.org/10.1111/cdev.12512} {A {{New View}} of
  {{Language Development}}: {{The Acquisition}} of {{Lexical Tone}}}.
\newblock \emph{Child Development}, 87(3):834--854.

\bibitem[{Singh et~al.(2015)Singh, Goh, and
  Wewalaarachchi}]{singhSpokenWordRecognition2015}
Leher Singh, Hwee~Hwee Goh, and Thilanga~D. Wewalaarachchi. 2015.
\newblock \href {https://doi.org/10.1016/j.cognition.2015.05.010} {Spoken word
  recognition in early childhood: {{Comparative}} effects of vowel, consonant
  and lexical tone variation}.
\newblock \emph{Cognition}, 142:1--11.

\bibitem[{So and Best(2010)}]{soCrosslanguagePerceptionNonnative2010}
Connie~K. So and Catherine~T. Best. 2010.
\newblock Cross-language {{Perception}} of {{Non-native Tonal Contrasts}}:
  {{Effects}} of {{Native Phonological}} and {{Phonetic Influences}}.
\newblock \emph{Language and speech}, 53(Pt 2):273--293.

\bibitem[{Tsukada and Kondo(2019)}]{tsukadaPerceptionMandarinLexical2019}
Kimiko Tsukada and Mariko Kondo. 2019.
\newblock \href {https://doi.org/10.1177/0023830918806550} {The {{Perception}}
  of {{Mandarin Lexical Tones}} by {{Native Speakers}} of {{Burmese}}}.
\newblock \emph{Language and Speech}, 62(4):625--640.

\bibitem[{Vaswani et~al.(2017)Vaswani, Shazeer, Parmar, Uszkoreit, Jones,
  Gomez, Kaiser, and Polosukhin}]{vaswaniAttentionAllYou2017}
Ashish Vaswani, Noam Shazeer, Niki Parmar, Jakob Uszkoreit, Llion Jones,
  Aidan~N. Gomez, Lukasz Kaiser, and Illia Polosukhin. 2017.
\newblock \href {http://arxiv.org/abs/1706.03762} {Attention {{Is All You
  Need}}}.

\bibitem[{Wang and Zhang(2015)}]{wangTHCHS30FreeChinese2015}
Dong Wang and Xuewei Zhang. 2015.
\newblock \href {http://arxiv.org/abs/1512.01882} {{{THCHS-30}} : {{A Free
  Chinese Speech Corpus}}}.

\bibitem[{Wang and Chen(2020)}]{wangAcquisitionMandarinConsonants2020}
Xinchun Wang and Jidong Chen. 2020.
\newblock \href {https://doi.org/10.3390/languages5020020} {The {{Acquisition}}
  of {{Mandarin Consonants}} by {{English Learners}}: {{The Relationship}}
  between {{Perception}} and {{Production}}}.
\newblock \emph{Languages}, 5(2):20.

\bibitem[{Wilcox et~al.(2023)Wilcox, Futrell, and
  Levy}]{wilcoxUsingComputationalModels2023}
Ethan~Gotlieb Wilcox, Richard Futrell, and Roger Levy. 2023.
\newblock \href {https://doi.org/10.1162/ling_a_00491} {Using {{Computational
  Models}} to {{Test Syntactic Learnability}}}.
\newblock \emph{Linguistic Inquiry}, pages 1--44.

\bibitem[{Wu et~al.(2020)Wu, {Adda-Decker}, and Lamel}]{Wu2020MandarinDuration}
Yaru Wu, Martine {Adda-Decker}, and Lori Lamel. 2020.
\newblock \href {https://doi.org/10.21437/Interspeech.2020-1614{\"{i}}}
  {Mandarin lexical tones: {{A}} corpus-based study of word length, syllable
  position and prosodic position on duration}.
\newblock pages 1908--1912.

\bibitem[{Xi et~al.(2009)Xi, JIANG, Zhang, and
  Shu}]{xiCategoricalPerceptionVOT2009}
Jie Xi, Wei JIANG, Linjun Zhang, and Hua Shu. 2009.
\newblock \href {https://doi.org/10.3724/SP.J.1041.2009.00572} {Categorical
  {{Perception}} of {{VOT}} and {{Lexical Tones}} in {{Chinese}} and the
  {{Developmental Course}}}.
\newblock \emph{Acta Psychologica Sinica}, 41:572--579.

\bibitem[{Yeung et~al.(2013)Yeung, Chen, and Werker}]{yeungWhenDoesNative2013}
H.~Henny Yeung, Ke~Heng Chen, and Janet~F. Werker. 2013.
\newblock When does native language input affect phonetic perception? {{The}}
  precocious case of lexical tone.
\newblock \emph{Journal of memory and language}, 68(2):123--139.

\bibitem[{Yip(2002)}]{yipTone2002}
Moira Yip. 2002.
\newblock \href {https://doi.org/10.1017/CBO9781139164559} {\emph{Tone}}.
\newblock Cambridge {{Textbooks}} in {{Linguistics}}. {Cambridge University
  Press}, {Cambridge}.

\bibitem[{Yuan et~al.(2021)Yuan, Ryant, Cai, Church, and
  Liberman}]{yuanAutomaticRecognitionSuprasegmentals2021}
Jiahong Yuan, Neville Ryant, Xingyu Cai, Kenneth Church, and Mark Liberman.
  2021.
\newblock \href {http://arxiv.org/abs/2108.01122} {Automatic recognition of
  suprasegmentals in speech}.

\bibitem[{Zee(1991)}]{zeeChineseHongKong1991}
Eric Zee. 1991.
\newblock \href {https://doi.org/10.1017/S0025100300006058} {Chinese ({{Hong
  Kong Cantonese}})}.
\newblock \emph{Journal of the International Phonetic Association},
  21(1):46--48.

\bibitem[{Zhu et~al.(2022)Zhu, Zhang, and Jurgens}]{zhu2022charsiu}
Jian Zhu, Cong Zhang, and David Jurgens. 2022.
\newblock Phone-to-audio alignment without text: {{A Semi-supervised
  Approach}}.
\newblock \emph{IEEE International Conference on Acoustics, Speech and Signal
  Processing (ICASSP)}.

\end{thebibliography}

\end{document}